\documentclass{article}

\usepackage{arxiv}

\usepackage[utf8]{inputenc} 
\usepackage[T1]{fontenc}    
\usepackage{hyperref}       
\usepackage{url}            
\usepackage{booktabs}       
\usepackage{amsfonts}       
\usepackage{nicefrac}       
\usepackage{microtype}      
\usepackage{lipsum}		
\usepackage{graphicx}
\usepackage{natbib}
\usepackage{doi}
\usepackage{amsmath, amsthm, amssymb}
\usepackage{booktabs}
\usepackage{multirow}
\usepackage{tabularx}
\usepackage{bbm}
\usepackage{bm}
\usepackage{subfigure}
\usepackage{caption} 
\usepackage{soul}
\usepackage{makecell}
\usepackage{xcolor}
\usepackage{soul}
\newcommand{\nguyen}[1]{{\color{black}{#1}}}
\newcommand{\dung}[1]{{\color{black}{#1}}}
\newcommand{\fix}[1]{{\color{black}{#1}}}

\usepackage{subfigure}
\usepackage{caption} 
\usepackage{graphicx} 
\newtheorem{prop}{Proposition}

\newtheorem*{assumption*}{Assumption}
\newtheorem{observation}{Observation}
\newtheorem{mydef}{Definition}
\usepackage[linesnumbered, ruled,vlined]{algorithm2e}
\usepackage{algorithmic}
\SetAlFnt{\footnotesize}
\SetKw{KwDownTo}{downto}
\SetKw{KwTo}{to}
\SetKw{KwTrue}{true}
\SetKw{KwFalse}{false}
\SetKwInOut{Input}{Input}
\SetKwInOut{Output}{Output}
\SetKw{KwAnd}{and}
\SetKw{KwBreak}{break}

\usepackage{amsthm}

\usepackage{booktabs} 
\usepackage{multirow}
\usepackage{tabularx}
\newtheorem{theorem}{Theorem}

\title{FedGrad: Mitigating Backdoor Attacks in Federated Learning Through Local Ultimate Gradients Inspection}

\author{ 
{
Thuy Dung Nguyen} \\
	College of Engineering \& Computer Science\\
        VinUni-Illinois Smart Health Center\\
	VinUniversity, Vietnam \\
	\texttt{dung.nt2@vinuni.edu.vn} \\
	\And
{
 Anh Duy Nguyen} \\
	College of Engineering \& Computer Science\\
        VinUni-Illinois Smart Health Center\\
	VinUniversity, Vietnam \\
	\texttt{duy.na@vinuni.edu.vn} \\
 \AND
	Kok-Seng Wong \\
        College of Engineering \& Computer Science\\
        VinUni-Illinois Smart Health Center\\
        VinUniversity, Vietnam \\
	\texttt{wong.ks@vinuni.edu.vn} \\
 \And
	Huy Hieu Pham \\
        College of Engineering \& Computer Science\\
        VinUni-Illinois Smart Health Center\\
        VinUniversity, Vietnam \\
	\texttt{hieu.ph@vinuni.edu.vn} \\
 \AND
 	Thanh Hung Nguyen \\
        Hanoi University of Science and Technology\\
        Hanoi, Vietnam \\
	\texttt{hungnt@soict.hust.edu.vn} \\
 \And 
  	Phi Le Nguyen \thanks{Corresponding authors}\\
        Hanoi University of Science and Technology\\
        Hanoi, Vietnam \\
	\texttt{lenp@soict.hust.edu.vn} \\
 \AND 
        $\text{Truong Thao Nguyen }^*$\\
        The National Institute of Advanced Industrial Science \\
        and Technology (AIST), Japan \\
	\texttt{nguyen.truong@aist.go.jp} \\
}

\date{}

\renewcommand{\shorttitle}{FedGrad}

\hypersetup{
pdftitle={FedGrad: Mitigating Backdoor Attacks in Federated Learning Through Local Ultimate Gradients Inspection},
pdfauthor={Thuy Dung Nguyen, Anh Duy Nguyen, Thanh-Hung Nguyen, Kok-Seng Wong, Huy Hieu Pham, Truong Thao Nguyen, Phi Le Nguyen},
}

\begin{document}
\maketitle

\begin{abstract}
    Federated learning (FL) enables multiple clients to train a model without compromising sensitive data. The decentralized nature of FL makes it susceptible to adversarial attacks, especially backdoor insertion during training. Recently, the edge-case backdoor attack employing the tail of the data distribution has been proposed as a powerful one, raising questions about the shortfall in current defenses’ robustness guarantees. Specifically, most existing defenses cannot eliminate edge-case backdoor attacks or suffer from a trade-off between backdoor-defending effectiveness and overall performance on the primary task. To tackle this challenge, we propose FedGrad, a novel backdoor-resistant defense for FL that is resistant to cutting-edge backdoor attacks, including the edge-case attack, and performs effectively under heterogeneous client data and a large number of compromised clients. FedGrad is designed as a two-layer filtering mechanism that thoroughly analyzes the ultimate layer’s gradient to identify suspicious local updates and remove them from the aggregation process. We evaluate FedGrad under different attack scenarios and show that it significantly outperforms state-of-the-art defense mechanisms. Notably, FedGrad can almost 100\% correctly detect the malicious participants, thus providing a significant reduction in the backdoor effect (e.g., backdoor accuracy is less than 8\%) while not reducing main accuracy on the primary task.
\end{abstract}

\section{Introduction}
\label{sec:intro}
\textbf{Background.} Federated Learning (FL) emerges as a promising solution that enables the use of data and computing resources from multiple clients to train a shared model under the orchestration of a central server \cite{federated-learning}. In FL, clients use their own data to train the model locally and iteratively share the local updates with the server, which then combines the contributions of the participating clients to generate a global update. 
Although FL has many notable characteristics and has been successful in many applications~\cite{mobile-edge}, recent studies indicate that FL is fundamentally susceptible to adversarial attacks, in which malicious clients manipulate the local training process to contaminate the global model~\cite{pmlr-v108-bagdasaryan20a}. 
Based on the goal of the attack, adversarial attacks can be broadly classified into untargeted \fix{and} targeted attacks.
The former aims to deteriorate the performance of the global model on all test samples~\cite{krum, gaussian_MP}, while the latter (also known as a backdoor attack) focuses on causing the model to generate false predictions for adversary-chosen inputs~\cite{Sun2019CanYR,pmlr-v108-bagdasaryan20a}.
Recently, \cite{NEURIPS2020_b8ffa41d} proposed and proved the stealthiness of \textit{edge-case backdoor attacks}, which raises a great concern for the current defense schemes. In an edge-case attack, 
the adversary targets input data points that are far from the \fix{global} distribution and unlikely to appear on the validation set of benign clients' training data. \\

\textbf{Existing approaches and their limitations.} Many efforts have been devoted to dealing with existing threats in FL, which can be roughly classified into two directions: {robust FL-aggregation and anomaly model detection}. The former aims to optimize the aggregation function in order to limit the effects of polluted updates caused by attackers, whereas the latter attempts to identify and remove malicious updates. The most common way to lessen malicious updates is by restricting all clients' updates by a threshold. For instance, \cite{pmlr-v139-xie21a} presented a certified defense mechanism based on the clipping and perturbation paradigm.
Other approaches focused on new estimators such as coordinate-wise median, $\alpha$-trimmed mean ~\cite{pmlr-v80-yin18a}, geometric median~\cite{rfa} for aggregation. 
The main drawback of the aforementioned methods is that polluted updates remain in the global model, thereby reducing the model's precision {while not totally mitigating the backdoor's impact}. In addition, clipping the update norm of all clients (including benign clients) will lessen the magnitude of the updates and slow down the model's convergence speed.

Several methods have been proposed to identify adversarial clients and remove them from the aggregation. In~\cite{Auror}, the authors proposed a defense mechanism against poisoning attacks in collaborative learning based on \fix{the} $K$-Means algorithm.
~\cite{Sattler2020OnTB-cluster} proposed dividing the clients' updates into normal updates and suspicious updates based on their cosine similarities. 
In the aforementioned defense schemes, the entire information from clients' updates is used to measure client similarity. However, only a tiny portion of updates have a significant impact on the backdoor task, while most are negligible~\cite{Zhou2021DeepMP}. Recognizing this fact,~\cite{Rieger2022DeepSightMB} proposed identifying malicious clients by analyzing the internal structure of the last layer's weights \fix{and comparing} the outputs of the clients' local models. Nonetheless, this strategy \fix{demands the creation of extra validation sets} and local inference for each client, raising serious questions about their applicability in large-scale FL systems.

Additionally, most of the methods for identifying malicious clients proposed so far follow the majority-based paradigm in the sense that they assume benign local model updates are a majority compared to the malicious ones; thus, polluted updates are supposed to be outliers in the distribution of all updates.
Unfortunately, this hypothesis holds true only if the data of the clients is IID (independent and identically distributed) and the number of malicious clients is small. 
In many real-world scenarios, client data exhibits a high level of heterogeneity, and the number of adversarial clients in a communication round can be substantial. This fact has 
imposed a great challenge on the backdoor attack resistance, making it an open issue \cite{NEURIPS2020_b8ffa41d}.\\

\textbf{Our solution.} In this paper, we propose a defense mechanism named FedGrad, capable of mitigating backdoor attacks under the most stressful environments, i.e., in the presence of heterogeneous client data and a significant number of adversarial clients.
The proposed approach is based on \fix{two} significant findings: 1) the discriminative capability of the last layer's gradient (or ultimate gradient for short) and 2) the dynamics of clustering characteristics of clients’ local updates. 
Firstly, through comprehensive theoretical analysis and empirical experiments, we discovered that the ultimate gradient conveys rich information about local training objectives, allowing us to accurately distinguish malicious and benign participants.
Secondly, we have \fix{the essential observation} that the clustering feature of clients' local updates varies throughout communication rounds. 
Specifically, in the initial few rounds, the benign client models show a significant degree of diversity, while the poisoned models may exhibit a high degree of similarity as they are trained with the same attack objective.
In contrast, as the model is trained increasingly, the local models of benign clients tend to converge on the same global model, bringing them closer together.

In light of these findings, we propose a two-layer ultimate gradient-based filtering technique to identify suspicious model updates and eliminate them from the aggregated model, rather than using a unique clustering algorithm to distinguish between malicious and benign clients. Each layer in our proposed method is responsible for dealing with a specific characteristic of the client updates.  
Our proposed filtering algorithm can perform effectively even \fix{in} the presence of heterogeneous client data and a large number of adversarial clients.
The work that is the most relevant to our approach is DeepSight~\cite{Rieger2022DeepSightMB}.
Unlike DeepSight, however, our clustering approach is based purely on the ultimate gradient without the requirement to perform inference on local updates of all clients during each communication round, thereby avoiding the additional computing overhead and ensuring scalability.
\fix{Furthermore, unlike DeepSight's clustering technique, our suggested two-layer filtering mechanism adapts appropriately to the fluctuating clustering features of client updates, enhancing its ability to filter malicious clients.} These points \fix{make} our proposed method superior to DeepSight. 

In summary, our contributions are specified as follows:\\[-0.1cm]
\begin{itemize}
    \item We introduce FedGrad -- a novel backdoor-resistant defense for FL. It identifies and eliminates suspicious clients that distort the global model. FedGrad is the first work that, to the best of our knowledge, thoroughly addresses not only edge-case \fix{attacks} but also other common types of backdoor attacks, even in the presence of a large number of compromised participants and a high degree of client data heterogeneity.\\[-0.1cm]
    \item We perform thorough, comprehensive theoretical analysis and empirical experiments on the gradient of the last layer and achieve significant findings regarding the discriminative capability of the last layer’s gradient and the dynamics of clustering characteristics of clients’ local updates.\\[-0.1cm]
    \item We design a robust ultimate-gradient-based two-layer filtering mechanism that accurately \fix{identifies} malicious participants. The two filter layers take advantage of the discriminative capability of the last layer’s gradient  and possess complementary features to accommodate the clustering dynamics of clients' local updates. We also leverage the historical relationship between participants from the previous communication
    rounds to make anomaly detection in FedGrad more reliable.\\[-0.1cm]
    \item We conduct comprehensive experiments and in-depth ablation \fix{studies} on various datasets, models, and backdoor attack scenarios to demonstrate the superiority of FedGrad over state-of-the-art defense techniques.
\end{itemize}

\section{Threat Model}
\label{sec:preliminaries}

\begin{figure*}[tb]
    \centering
\begin{minipage}{0.24\linewidth}
\includegraphics[width=\columnwidth]{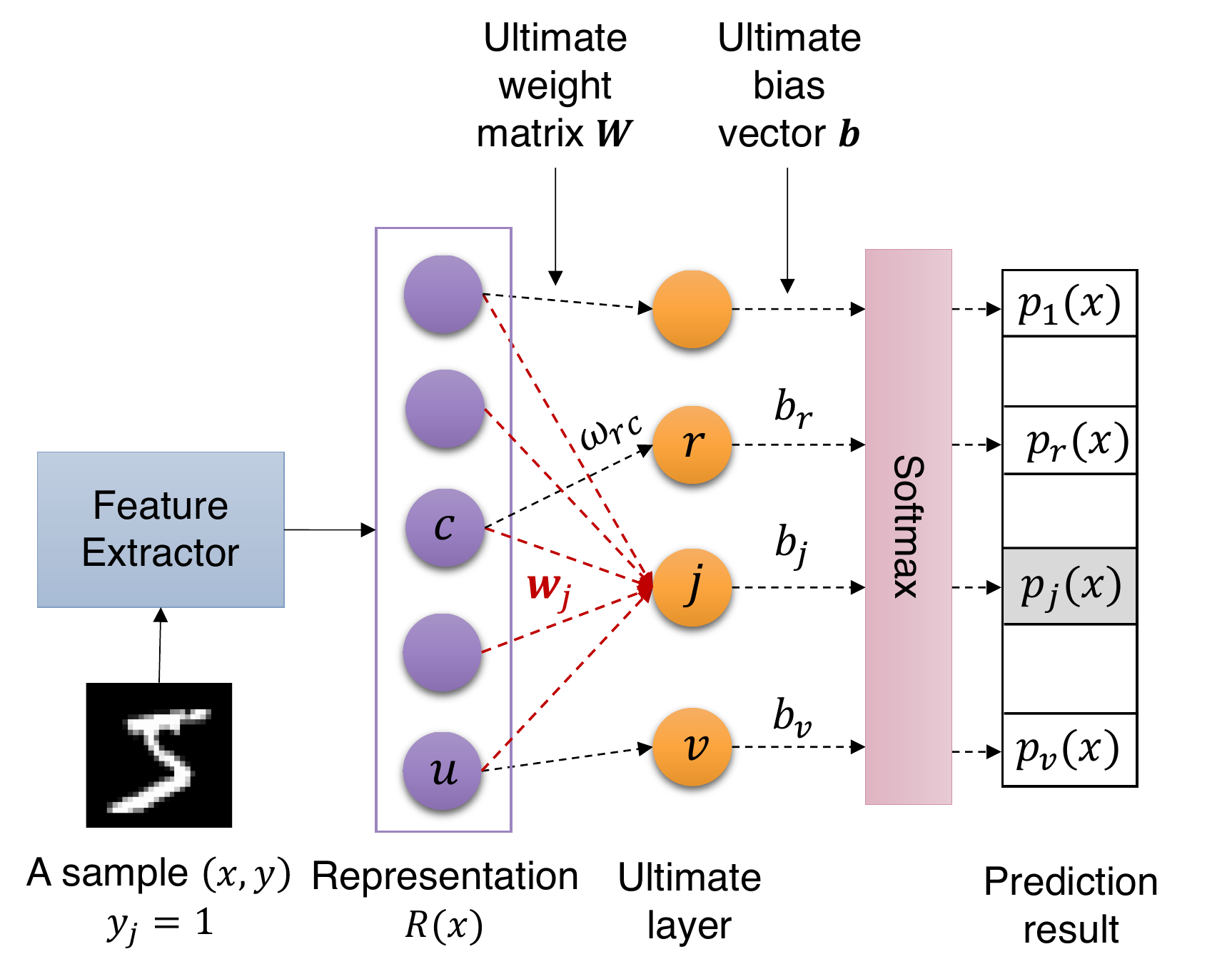}
    \caption{\footnotesize \textbf{Illustration of the ultimate layer and the ultimate gradient.} The ultimate layer is the last layer connected to the penultimate layer via the ultimate weight matrix.} 
    \label{fig:penultimate}
\end{minipage}
\hfill
\begin{minipage}{0.43\linewidth}
    \centering
    \includegraphics[width=0.95\columnwidth]{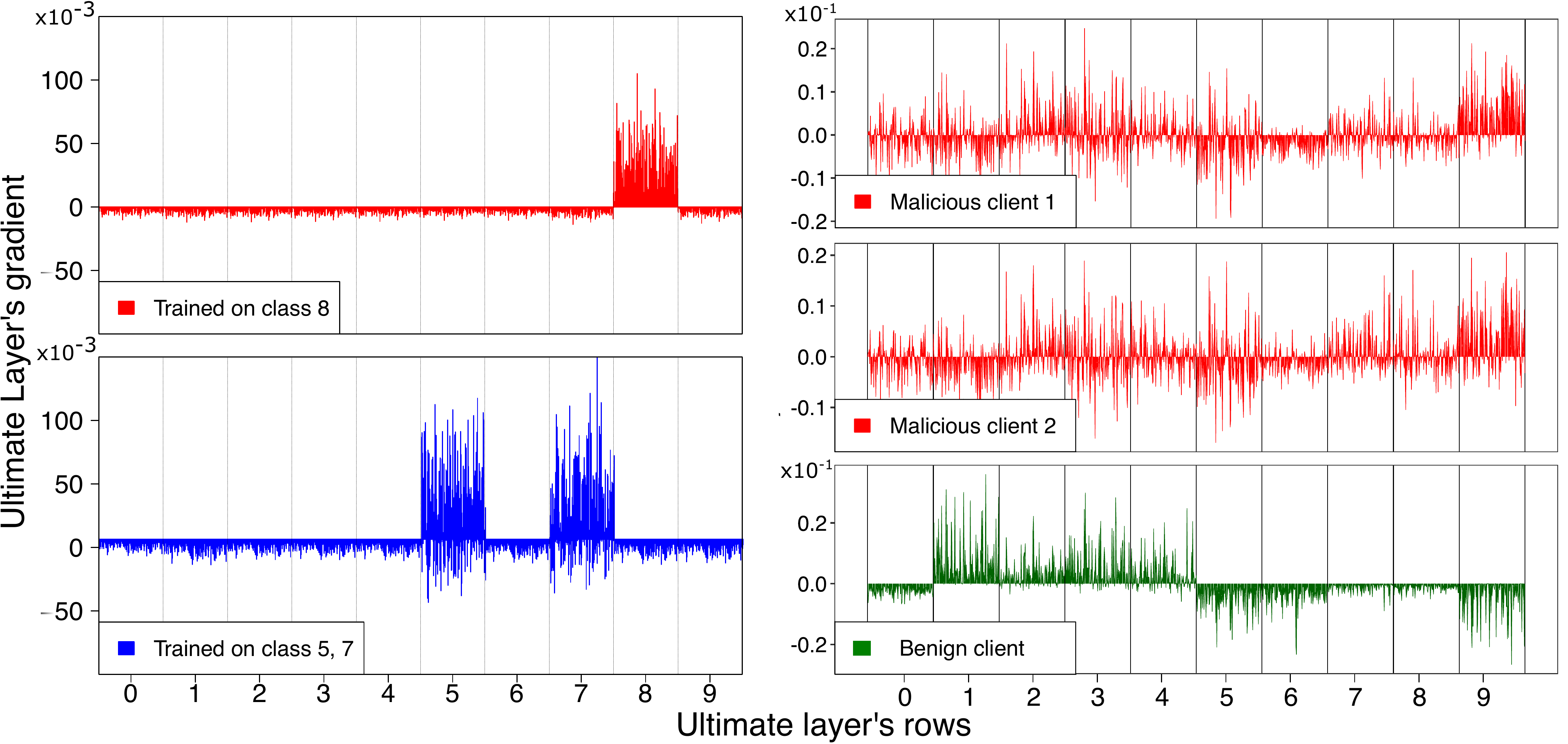}
   \caption{\footnotesize \textbf{The behavior of ultimate gradient.}  
   When training with samples from specific classes (the left sub-figure), the rows corresponding with the trained classes may contain positive elements, while the other rows \fix{are} strictly negative. 
   As the adversarial clients share the same training objective, their ultimate gradients (\fix{the} red color in the right sub-figure) have similar patterns. \label{penultimate_layer_gradient}}
\end{minipage}
\hfill
\begin{minipage}{0.29\linewidth}
    \centering
    \includegraphics[width=0.95\columnwidth]{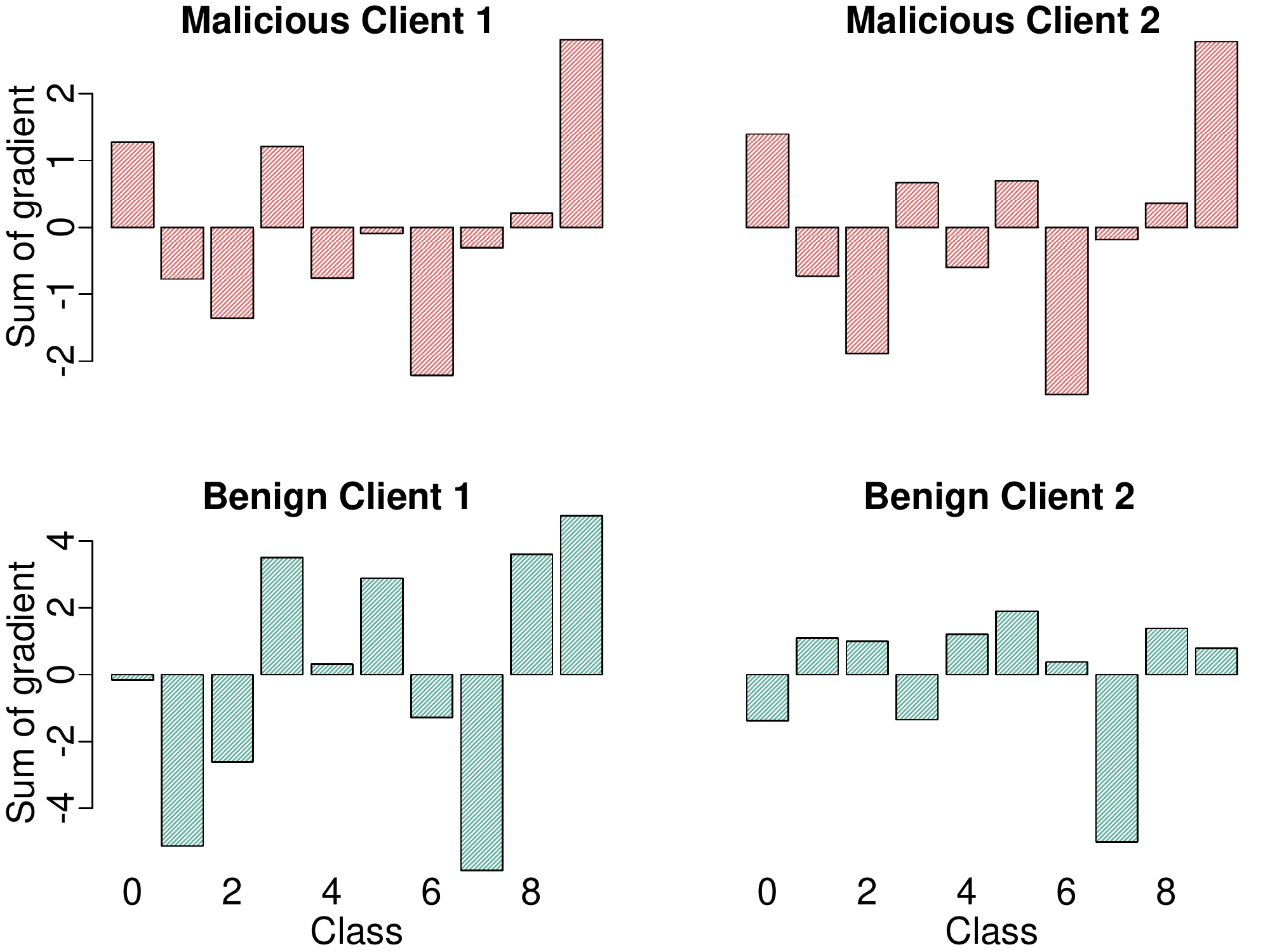}
    \caption{\footnotesize \textbf{Discriminative characteristic of by-class ultimate weight gradient sum.} \footnotesize 
    Malicious updates (in red color) share the same pattern due to similar training objectives, while those of benign updates differ significantly (non-IID distribution).}
    \label{fig:penultimate_grad}
\end{minipage}
\vspace{-0.3cm}
\end{figure*}
FL gives the adversary $\mathcal{A}$ complete control over a set $\mathcal{C}$ of compromised clients in order to obtain poisoned models, which are later aggregated into the global model $G$ to poison its properties. 
In particular, $\mathcal{A}$ wants the poisoned model $G'$ to behave normally on all inputs except for specific inputs $x \in \mathcal{T}_{\mathcal{A}}$ (where $\mathcal{T}_{\mathcal{A}}$ denotes the so-called targeted class), for which attacker-chosen (incorrect) predictions should be output. 
We denote $\epsilon$ as the compromising ratio, i.e., $\epsilon = \frac{|\mathcal{C}|}{N}$, where $N$ is the total number of clients in the FL system. In each round, the aggregation server will randomly choose $K$ clients from the total $N$ clients\fix{; thus,} the number of malicious participants in each training round ranges from $0$ to $\min\left(\epsilon \cdot N, K\right)$.
The attacker's capabilities can be summarized as follows: (1) the attacker manipulates the local training data of any compromised client; (2) it completely controls the local training procedure and the hyperparameters (i.e., number of epochs); and (3) it modifies the local update of each resulting model before submitting it for aggregation. In addition, $\mathcal{A}$ is knowledgeable of the server's operations. However, $\mathcal{A}$ has no control over any processes executed at the server, nor over the benign clients. 
In the following, we will briefly describe the techniques that the adversary $\mathcal{A}$ uses to achieve its objectives.\\
\textbf{Data poisoning.} $\mathcal{A}$ adds a poisoned sample set $\mathcal{D}_{p}$ to the training dataset of each client $i$ in the compromising set $\mathcal{C}$.
In which, training samples of each compromised client $i$ are sampled from the same distribution.
We denote the fraction of injected poisoned data $\mathcal{D}_{p}$ in the overall poisoned training dataset $\mathcal{D}'_i$ of client $i$ as Poisoned Data Rate (PDR), i.e.,
$PDR = \frac{|\mathcal{D}_{p}|}{|\mathcal{D}^{'}_i|}$. \\
\textbf{Model replacement (MR).} 
After the local training procedure is completed on the poisoned data (i.e., $\mathcal{D}'_i$), $\mathcal{A}$ intensifies the backdoor effect by scaling the model updates before submitting \fix{them} back to the server~\cite{pmlr-v108-bagdasaryan20a}. 
To make the attack more stealthy, projected gradient descent (PGD) ~\cite{Sun2019CanYR} is applied to ensure that each poisoned model does not differ considerably from the global model at each communication round. 
PGD is often combined with model replacement to achieve better effectiveness \cite{NEURIPS2020_b8ffa41d}.

\section{Gradient of Ultimate Layer}
\label{ssec:penultimate_gradient}
Let consider a typical deep neural network $\mathcal{M}$ for a classification task, consisting of a feature extractor and a classifier that is trained using the cross entropy loss by SGD method. 
We assume that the classifier comprises of a dense layer, followed by a softmax, and a bias vector layer 
(Fig.~\ref{fig:penultimate}). 

\begin{mydef}[\textbf{Ultimate gradient}]
The ultimate layer is the last fully connected layer. 
Let $\textbf{W}$ denote the weight matrix connecting the ultimate layer and the previous one, and $\textbf{b}$ denote the last bias vector. Then, the ultimate gradient is the improvement of $\textbf{W}$ and   $\textbf{b}$ obtained after training the model $\mathcal{M}$ with the SGD method. 
Specifically, let $\textbf{W}^t$ and $\textbf{b}^t$ be the value of the $\textbf{W}$ and $\textbf{b}$ at the $t$-th training iteration, then the ultimate gradient at the $t$-th iteration $\textbf{g}^t$ is the combination of ultimate weight gradient $ \nabla \textbf{W}^t=\frac{\textbf{W}^t - \textbf{W}^{t-1}}{-\eta}$, and ultimate bias gradient $\nabla \textbf{b}^t=\frac{\textbf{b}^t - \textbf{b}^{t-1}}{-\eta}$,
where $\eta$ is the learning rate.
\end{mydef}
To ease the presentation, 
we may omit the index $t$ and simply represent the ultimate gradients by $\textit{\textbf{g, W, b}}$.
The number of the ultimate layer's neurons equals to the number of classes, with the $j$-th neuron indicating the prediction result for the $j$-th class. 
Intuitively, the $j$-th row of the ultimate weight matrix and the $j$-th element of the bias vector contain the most representative information of the training samples of the $j$-th class.
In the following, we introduce a new term named \emph{by-class ultimate gradient sum} to highlight the effect of the training data on the ultimate layer. 
\begin{mydef}[\textbf{By-class ultimate gradient sum}]
The by-class ultimate gradient sum, denoted by $\mu$, is obtained by summarizing the ultimate gradient by class, i.e., $\mu = \left<\textbf{g} \cdot \mathbbm{1}\right> {=} \left<\nabla \textbf{W} \cdot \mathbbm{1}, \nabla \textbf{b}\right>$. Note that $\mu$ consists of two $v$-dimensional vectors, with the first representing the by-row sum of the ultimate weight gradient matrix and the second representing the ultimate bias gradient vector. 
\end{mydef}

\begin{prop}[\textbf{Ultimate gradient when training with a single class}]
\label{prop:pen_shift}
Suppose $\textbf{W}{=}[\textbf{w}_1, ..., \textbf{w}_v]$, and $\textbf{b} {=} \left < b_1, ..., b_v \right > $, where $\textbf{w}_i$ is the $i$-th row of $W$, and $b_i$ is the $i$-th element of $\textbf{b}$. Let $(x, y)$ be a sample with the groundtruth label of $j$, i.e., $(y_j = 1, y_i = 0$ $\forall i \neq j)$.  
After training the model $\mathcal{M}$ with sample $(x,y)$,
the values of all elements in $\textbf{w}_j$ and $b_j$ have an inverse sign compared with other elements in $\textbf{W}$ and $\textbf{b}$.
\end{prop}
\noindent \textbf{Proof.} Let $x$ be a training sample, and $R = R(x) \in \mathbb{R}^u$ denote the representation of $x$. 
Let us denote by $L(x) \in \mathbb{R}^v$  the logits of $x$, then $L(x)$ is defined as
\begin{equation}
\small
\nonumber
    L(x) = \textbf{W} \cdot R + \mathbf{b} = \begin{bmatrix}
      R_1 w_{11} + R_2 w_{12} + \dotsc + R_u w_{1u} + b_1           \\[0.3em]
      R_1 w_{21} + R_2 w_{22} + \dotsc + R_u w_{2u} + b_2           \\[0.3em]
      \vdots\\[0.3em]
      R_1 w_{v1} + R_2 w_{v2} + \dotsc + R_u w_{vu} + b_v
     \end{bmatrix}.
\end{equation}
Let $p(x)$ be the prediction result, which is the output of the softmax layer, then the probability of sample $x$ being classified into class $j$, i.e., $p_j(x)$, is determined as 
\begin{equation}
\small
    p_j(x) = \frac{e^{L_j}}{\sum_{i=1}^{v} e^{L_i}}.
\end{equation}
The cross entropy loss $\mathcal{L}(p(x), y)$  concerning sample $(x,y)$ is given by
\begin{equation}
\small
    \mathcal{L}(p(x), y) = \sum_{i=0}^{v} {y_i \log \bigg(\frac{1}{p_i(x)}\bigg)}.
\end{equation}
Let $w_{rc}$ be the item at row $r$ and column $c$ of $W$, then the gradient of the loss $\mathcal{L}(p(x), y)$ with respect to $w_{rc} \in W$ is 

\begin{equation}
    \small
    \frac{\partial \mathcal{L}}{\partial w_{rc}} = \sum_{i=1}^{v}{\bigg(\frac{\partial \mathcal{L}}{\partial p_i(x)} \cdot 
    \bigg(\sum_{k=1}^{v}\frac{\partial p_i(x)}{\partial L_k} \cdot \frac{\partial L_k}{\partial w_{rc}} \bigg)\bigg)}. 
    \label{eqn:grad}
\end{equation}
\noindent We have
\begin{equation}
    \small
    \frac{\partial \mathcal{L}}{\partial p_i(x)} = -\frac{1}{\ln{10}}\frac{y_i}{p_i(x)} = 
    \begin{cases}
        -\frac{1}{\ln{10}}\frac{y_j}{p_j(x)} & \text{if $i=j$}\\
        0 & \text{if $i\neq j$}
    \end{cases},
    \label{eqn:grad1}
\end{equation}

\begin{align}
    \small
    \label{eqn:grad2} \frac{\partial p_j(x)}{\partial L_k} &= 
    \begin{cases}
        p_j(x)(1-p_k(x)), & \text{if $k=j$} \\
        -p_j(x) p_k(x), & \text{otherwise}
    \end{cases};\\
    \frac{\partial L_k}{\partial w_{rc}} & = 
    \begin{cases}
        R_c, & \text{if $k=r$} \\
        0, & \text{otherwise}
    \end{cases}; 
    \label{eqn:grad3}
\end{align}

\noindent From (\ref{eqn:grad1}), (\ref{eqn:grad2}) and (\ref{eqn:grad3}), we deduce that 
\begin{equation}
    \label{eq:gradient}
    \frac{\partial \mathcal{L}}{\partial w_{rc}} = 
    \begin{cases}
        \frac{-1}{\ln{10}}y_j (1 - p_j(x)) R_c, & \text{if $r = j$},\\
        \frac{1}{\ln{10}}y_j p_r(x) R_c, & \text{otherwise.}
    \end{cases}
\end{equation}
\noindent Let $b_{r}$ be the bias at row $r$ of $W$, then the gradient of the loss $\mathcal{L}(p(x), y)$ with respect to $b_{r} \in W$ is 
\begin{align}
    \small
    \frac{\partial \mathcal{L}}{\partial b_{r}} &= \sum_{i=1}^{v}{\bigg(\frac{\partial \mathcal{L}}{\partial p_i(x)} \cdot 
    \bigg(\sum_{k=1}^{v}\frac{\partial p_i(x)}{\partial L_k} \cdot \frac{\partial L_k}{\partial b_{r}} \bigg)\bigg)} 
    \label{eqn:gradx} \\
    \label{eq:gradient2} & = \begin{cases}
        \frac{-1}{\ln{10}}y_j (1 - p_j(x)), & \text{if $r = j$},\\
        0, & \text{otherwise.}
    \end{cases}
\end{align}

\noindent 
As $y_j = 1$, $ 0 < p_i(x) < 1$ ($\forall i$), when applying the gradient descent, the values of the $j$-th row of $W$ increase/decrease while those on all other rows decrease/increase (depending on $R_c$ sign).
Analogously, the value of bias corresponding to the $j$-th row increases/decreases while those on all other rows follows the opposite trend.
Fig.~\ref{penultimate_layer_gradient} illustrates an intuition for Proposition \ref{prop:pen_shift}. 

\begin{observation}[\textbf{Ultimate gradient when training with datasets having similar distributions}]
\label{obs:grad_similar_dts}
Suppose $\mathcal{D}_1, \mathcal{D}_2, \ldots, \mathcal{D}_{\mathcal{C}}$ are $\mathcal{C}$ training datasets with the same distribution and $\mathcal{M}$ be a shared model. 
Let $\mathcal{M}_i$ ($i = 1, ..., \mathcal{C}$) be the model received by training $\mathcal{M}$ using $\mathcal{D}_i$. Then, the ultimate gradients of $\mathcal{M}_1, ..., \mathcal{M}_\mathcal{C}$ share the same pattern (i.e., distribution). 
\end{observation}
%
\noindent \textbf{Intuitive Proof.} Intuitively, all samples with the same label will likely result in similar representations after passing through the feature extractor.
Consequently, it can be deduced from Equations (\ref{eq:gradient}) and  (\ref{eq:gradient2}) that the ultimate gradients obtained after training with samples belonging to the same classes are similar.
As $\mathcal{D}_1, \mathcal{D}_2, \ldots, \mathcal{D}_K$ have the same distribution, training with them will produce models with similar ultimate gradients.

This observation suggests that if the data of benign clients is IID, their local models' ultimate gradients will exhibit similar patterns (Fig. \ref{penultimate_layer_gradient}); thus, the ultimate gradient can be used to distinguish between benign and malicious clients.
However, when the client data is non-IID, the situation becomes more challenging.
To this end, we have made the crucial observation that attacker typically uses a large poisoned data rate
to better achieve their attack objectives (cf. \cite{DBA}).

\begin{observation}[\textbf{Discriminative characteristic of ultimate gradient}]
In backdoor attacks, the by-class ultimate gradient sums of the compromised clients' models tend to have similar patterns. 
\end{observation}

\begin{figure}[t]
    \centering
\includegraphics[width=0.6\linewidth]{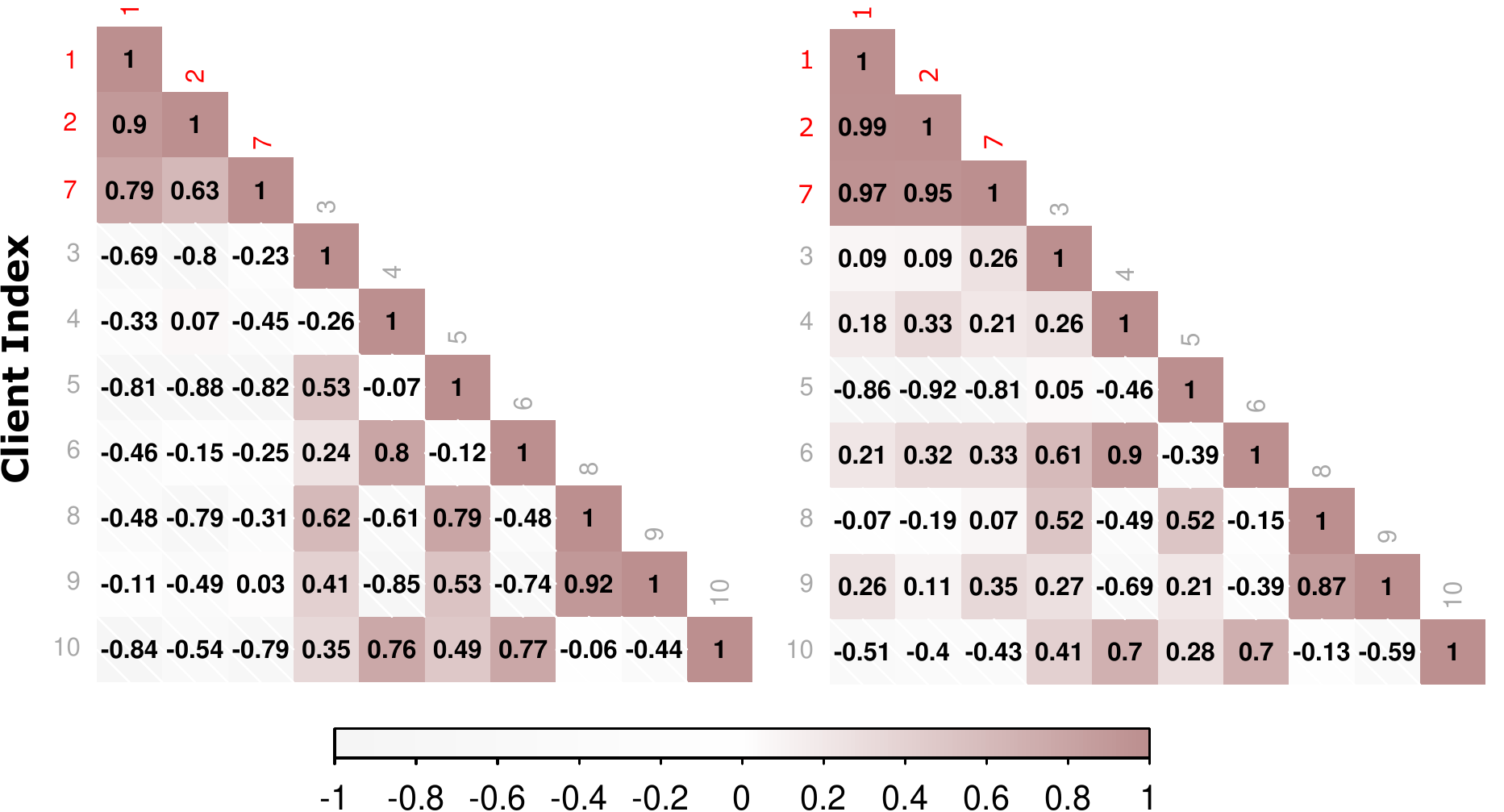}
        \caption{\footnotesize \textbf{Cosine similarity regarding by-class gradient sum.}
        \footnotesize The right matrix presents the pairwise similarities of 10 local models in \fix{terms} of by-class weight gradient sum, while the left side regards the by-class bias gradient. The malicious clients (1, 2, 7) have a high degree of similarity, whereas those of the benign clients are diverse.
        }
        \label{fig:pairwise_cs}
        \vspace{0.4cm}
\end{figure}

\noindent \textbf{Empirical justification.}
Fig. \ref{fig:penultimate_grad} plots the by-class gradient sums of both malicious and benign models. 
As observed, the by-class ultimate gradient sum of the two malicious clients follows a similar pattern since they use polluted data with the same distribution.
In contrast, the benign clients reveal different patterns as their training data is non-IID. 
To support our argument, we plot the cosine similarities of the clients' by-class ultimate gradient sum in Fig.~\ref{fig:pairwise_cs}. The left side of the figure depicts the similarity matrix for the by-class ultimate bias gradient, while the right side captures those concerning the by-class ultimate weight gradient sum. Obviously, the by-class ultimate gradient sums of the three malicious models are comparable, whereas the benign clients exhibit a great deal of variety.

\section{FedGrad: Against Backdoor Attacks via Ultimate Gradient}
\label{sec:FedGrad}
\begin{figure*}[tb]
    \centering
    \begin{minipage}{\linewidth}
            \centering
         \includegraphics[width=0.95\textwidth]{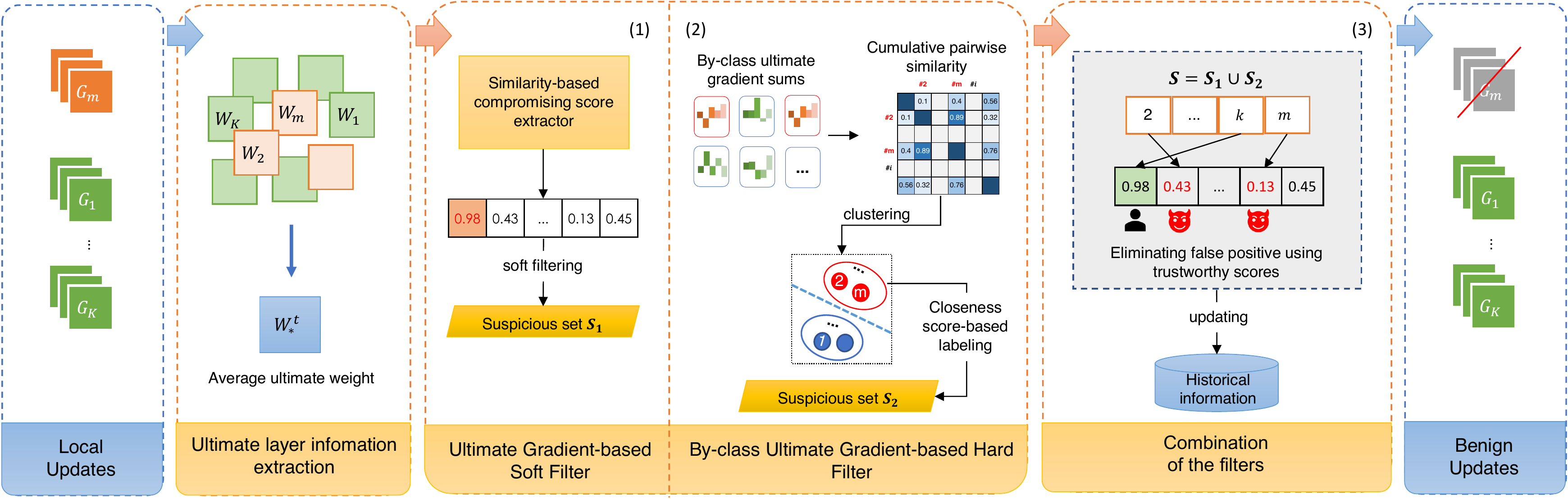}
        \caption{\footnotesize\textbf{Overview of FedGrad.} 
        Upon receiving local models from clients, the server: (1) applies the soft filter to determine the suspicious group $S_1$; (2) if the number of training rounds exceeds a threshold, then applies the hard filter to identify the suspicious group $S_2$; (3) combines $S_1$ and $S_2$ to obtain $S$, then utilizes the trustworthy score to remove the possibly benign clients from $S$; and finally, performs aggregation, excluding clients belonging to $S$.
        }
        \label{fig:fedgrad_detail}
    \end{minipage}
    \vspace{-0.3cm}
\end{figure*}
\subsection{Motivation and Overview of FedGrad} 
FedGrad is comprised of two filtering layers that are complementary to each other, and each layer has a particular role. 
In the initial few rounds, the benign client models show a significant degree of diversity, particularly when the client's data are non-IID.
In the meantime, as discussed in the previous section, poisoned models may exhibit a high degree of similarity as they are trained with the same attack objective and large amounts of polluted training data. 
Motivated by this observation, we designed the first filter, named \emph{soft filter}, based on the distance of clients' local models to a so-called average model (i.e., the one obtained by averaging all the local models). As the local models generated by the adversary are highly similar, their distances from the average model will be smaller than those of the benign models. Consequently, we may identify the corresponding poisoned updates by investigating the distance to the average model. 

However, as the model is trained increasingly, 
the local models of some benign clients may become closer to each other, i.e., when they converge to the same global model.
Therefore, the soft filter may fail. To address this issue, we propose a \emph{hard filter}, the second filter based on the clustering paradigm. In particular, we divide clients into two clusters using the by-class ultimate gradient sum and propose an algorithm for determining which cluster is benign.

Finally, to avoid misidentifying benign clients as malicious, we provide an additional mechanism to trace the clients' trustworthy score (estimating the likelihood for a client to be benign) and use this score to remove false positive results generated by the soft and hard filters.
The overall flow of the FedGrad (Fig. \ref{fig:fedgrad_detail}) is as follows: \\[-0.3cm]
\begin{enumerate}
    \item Upon receiving local updates, the aggregation server employs the soft filter to identify suspicious participants. Let us denote the set of these clients as $S_1$.\\[-0.3cm]
    \item If the number of training rounds exceeds a predefined threshold, the server employs the hard filter to divide the clients into two groups and identify the group that contains malicious clients. Let us refer to the determined malicious group as $S_2$.\\[-0.3cm]
    \item Let $S = S_1 \cup S_2$ 
    The server then eliminates from $S$ all clients whose trustworthy score exceeds a specified threshold. The remainder of $S$ is determined to be malicious clients. Finally, the server aggregates the updates from all clients, excluding those in $S$.
\end{enumerate}
\subsection{Ultimate Gradient-based Soft Filter}
\textbf{Compromising score calculation.} The gradient-based soft filter utilizes a dynamic threshold ($\epsilon_1$) to filter out a subset of suspicious clients in every training round. 
Let \textit{$\textbf{W}^{t}_1, \ldots, \textbf{W}^{t}_K$} be the ultimate weight matrices, and \textit{$\nabla \textbf{W}^{t}_1, \ldots, \nabla \textbf{W}^{t}_K$} be the ultimate weight gradients of the $K$ clients participating in the current training round $t$. 
Then, the average ultimate weight of the current round, denoted as \textit{$\textbf{W}^{t}_{*}$}, is defined by \textit{$\textbf{W}^{t}_{*}{=}\sum_{i=1}^{K}\frac{n_i}{n}\textbf{W}^{t}_i$}. 
Let \textit{$\nabla^{t}_{*}$} be the average ultimate weight gradient, which is defined by \textit{$\nabla^{t}_{*} {=} \sum_{i=1}^{K}\frac{n_i}{n}\nabla \textbf{W}^{t}_i$}. 
Then, the \emph{compromising score} of a client $i$ at round $t$, denoted by $cs^{t}_i$, is defined as \textit{$cs^{t}_i {=} \frac{\left(\textbf{W}^{t}_i-\textbf{W}^{t}_{*}\right)\left(\nabla^{t}_{*}\right)}{\|\textbf{W}^{t}_i-\textbf{W}^{t}_*\| \|\nabla^{t}_{*}\|}$}.
It can be intuitively shown that adversarial clients' compromising scores tend to be greater than those of benign clients.
Moreover, to emphasize the difference \fix{in} the compromising scores between the clients, we utilize the min-max scaling: $cs^{t}_i {\leftarrow} \frac{cs^{t}_i - \min_j{cs^{t}_j}}{\max_j{cs^{t}_j} - \min_j{cs^{t}_j}}$.\\
To enhance the reliability of the compromising score, rather than relying solely on the instant score of the current round, we also utilize the historical information from previous rounds to calculate the cumulative compromising score $cs_i$ as  $cs_i {\leftarrow} \frac{m-1}{m} cs_i + \frac{1}{m} cs^{t}_i$, where $m$ is the number of training rounds where client $i$ has joined the training process.

\textbf{Dynamic threshold determination.} 
To filter out malicious participants, we employ the adaptive threshold $\epsilon_1$ defined as $\epsilon_1 {=} \min ( \zeta,  \underset{i \in \mathcal{C}_t}{\text{median}} ~ cs_i )$, where $\mathcal{C}_t$ is the set of all clients participating in the current training round, and $\zeta$ is a fine-tuned hyper-parameter. 
The clients whose cumulative compromising score is greater than $\epsilon_1$ are determined to be malicious.
The rationale behind $\epsilon_1$ is as follows. 
First, it should be noted that the compromising score of malicious clients is typically higher than that of benign ones.
Therefore, when $median\left( cs_i \right)$ is moderately large, the number of adversarial clients participating in the current training round tends to be higher than usual.
As such, factor $\zeta$ accelerates the filter to reduce the number of missing malicious clients. In contrast, the moderately small $median\left( cs_i \right)$ infers that the number of malicious participants in this training round is small. Hence, the median is robust enough to separate out the malicious clients.

\subsection{By-class Ultimate Gradient-based Hard Filter}
\label{subsec:hardfilter}
The hard filter consists of two steps: (1) clustering participants into two groups, and (2) identifying the group of compromised clients.

\textbf{Clustering.} 
To ensure stability, we leverage the $K$-means algorithm to partition the clients into two distinct groups. The inputs for the $K$-means algorithm are a set of all clients' attribute vectors \textit{$\textbf{f}$}. Here, the attribute vector \textit{$\textbf{f}_i$} for each client $i$ is a vector representing the similarity of its by-class ultimate gradient to those of $K-1$ participants.
Specifically, \textit{$\textbf{f}_i = \left< w_{i1}, ..., w_{iK}; b_{i1}, ..., b_{iK} \right>$}, where $w_{ij}$ and $b_{ij}$ are the cosine similarities between the by-class ultimate weight gradient vectors and the by-class ultimate bias vectors of clients $i$ and $j$.
Given each client's attribute vector, the $K$-means algorithm is performed using the L2 distance.
Note that this kind of attribute helps to reflect not only the ultimate gradient of the clients but also their geometrical relation to the others, thereby enhancing the precision of our clustering algorithm.   

Moreover, to strengthen the reliability of the proposed similarity, we use historical data to cumulatively update $w_{ij}$ and $b_{ij}$ by $\left < w_{ij}, b_{ij} \right> = \frac{m-1}{m} \left< w_{ij}, b_{ij} \right> + \frac{1}{m}\left < w^t_{ij}, b^t_{ij} \right>$, where $\left< w_{ij}, b_{ij} \right>$ and $\left < w^t_{ij}, b^t_{ij} \right>$ are the cumulative and instant similarities of clients $i$ and $j$, $m$ is the number of rounds where both clients $i$ and $j$ join training together.

\textbf{Malicious group determination.}
Existing clustering-based defenses \cite{Sattler2020OnTB} typically identify the potentially malicious group as the one with the smallest number of clients.
Nonetheless, this is not always the case, particularly when the client's data is non-IID. 
Motivated by the method proposed in \cite{krum}, FedGrad uses a so-called \emph{closeness score} to identify the most likely benign clients.
The closeness score of a client $i$ is measured by the distance from its local model $G_i$ to its $K{-}f{-}2$ closest neighbors, where $f$ is the estimated number of adversarial clients in the current round. 
Then, the client most likely to be benign is the one with the smallest closeness score.
Given the benign client, we can easily identify the other group (that does not contain the benign client) as the malicious group.
The rationale behind our algorithm is that the benign clients' local models become closer to the global model when the global model has almost converged (cf. \cite{NEURIPS2020_b8ffa41d}). At the same time, the adversarial cluster presents its similarity within its group and has dissimilarity with the benign cluster. In addition, it is important to note that $K{-}f{-}2$ roughly corresponds to the number of benign clients, which is typically greater than the number of malicious clients.
Therefore, there should be some benign clients among a malicious client's $K{-}f{-}2$ closest neighbors. Consequently, the corresponding closeness score of a malicious client
tends to be greater than that of the benign client with the lowest closeness score.
\subsection{Filtering Strategy}
\textbf{Trustworthy score estimation.} 
Detecting and discarding malicious clients' updates from the aggregation is essential for preventing a poisoned global model.
However, mistaking honest clients for malicious ones could reduce the main task accuracy.
To this end, in addition to the two primary filters dedicated to identifying compromised clients, we utilize an additional filter to verify the results obtained from the two primary filters, thereby identifying false positives. We assign each participant in a training round an instant trustworthy score that reflects the likelihood that it is benign based solely on the information gathered from the current round.
\dung{The estimated trustworthy score for each participant is then derived by averaging the instant trustworthy scores throughout all training rounds, with false positives identified as those whose scores exceed a predefined threshold $\gamma$}. 
In particular, if a client is predicted to be malicious during a training round $t$, its instant trustworthy score is set to $\lambda_1$; otherwise, it is set to $\lambda_2 $. Here, $\lambda_1$ and $\lambda_2$ are two hyper-parameters.

\textbf{Combination of the filters.} 
Remind that the soft filter aims to figure out clients whose local models are close to the global one, while the hard filter focuses on clustering the clients into two distinct groups. 
In the initial rounds, client models are extremely diverse and do not exhibit clustering characteristics, so clustering-based methods are inappropriate at this stage. Therefore, in this phase, we employ only the gradient-based soft filter to determine the malicious participants. 
However, after several training rounds, the benign local models become more similar and show the same objectives as the global model, which means the soft filter may fail to correctly identify all the compromised clients. 
At this stage, the local models reveal a high degree of clustering. These facts necessitate the use of the hard filter for determining malicious group. 
Moreover, after using the soft and hard filters to roughly identify potential adversarial clients, the server leverages the trustworthy score to eliminate the false positive rate.

\section{Experiments}
\label{sec:evaluation}
\begin{table}[thb]
\caption{\nguyen{Trigger-based backdoor attacks setting.}}
\vspace{5pt}
\scriptsize
\centering
\setlength\tabcolsep{3pt} 

\resizebox{0.65\columnwidth}{!}{\begin{tabular}{@{}c|l|ccccc@{}}
\toprule
Attack Strategy& 
Dataset  & 
\begin{tabular}[]{@{}c@{}} PDR \end{tabular}&
\begin{tabular}[]{@{}c@{}}Scale \\factor \end{tabular}& 
\begin{tabular}[]{@{}c@{}}Compromising \\ratio $\epsilon$ \end{tabular}& 
\begin{tabular}[]{@{}c@{}}Benign lr \\/E \end{tabular}& 
\makecell{Poison lr \\/E} 
\\ 
\midrule
\multirow{2}{*}{\makecell{Unconstrained \\attack}} & MNIST    & 5/64         & 1          & 0.4               & 0.1/2        & 0.1/2                              \\
\cmidrule(r){2-7}
                                     & CIFAR-10 & 12/64         & 1           & 0.4               & 0.1/2        & 0.1/10                                      \\ \cmidrule(r){1-7}
\multirow{2}{*}{\makecell{Constrain \\and scale}}  & MNIST    & 5/64         & 100          & 0.25               & 0.1/2        & 0.1/2                                \\
\cmidrule(r){2-7}
                                     & CIFAR-10 & 6/25         & 20           & 0.25               & 0.1/2        & 0.1/10                         \\ \cmidrule(r){1-7}
\multirow{2}{*}{DBA}                 & MNIST    & 5/64         & 100          & 0.25               & 0.1/2        & 0.1/2                                    \\
\cmidrule(r){2-7}
                                     & CIFAR-10 & 6/25         & 20           & 0.25               & 0.1/2        & 0.1/10                                  \\ \bottomrule
\end{tabular}}
\label{tab:train_config_dba}
\end{table}

\begin{figure*}[!t]
\begin{minipage}{0.6\linewidth}
    	\centering
    	\small
    	\setlength\tabcolsep{3pt} 
    	\resizebox{\columnwidth}{!}{%
        \begin{tabular}{@{}l|ll|ll|ll||ll|ll|ll@{}}
        \toprule
        \multicolumn{1}{l|}{}
        & \multicolumn{6}{c||}{CIFAR-10} 
        & \multicolumn{6}{c}{EMNIST} 
        \\
        \multicolumn{1}{l|}{}
        & \multicolumn{2}{c}{Black-box}
        & \multicolumn{2}{c}{{PGD}}
        & \multicolumn{2}{c||}{{PGD + MR}}                     
        & \multicolumn{2}{c}{Black-box}
        & \multicolumn{2}{c}{{PGD}}
        & \multicolumn{2}{c}{{PGD + MR}}             \\
        \cmidrule(lr){2-3}\cmidrule(lr){4-5}\cmidrule(lr){6-7}
        \cmidrule(lr){8-9}\cmidrule(lr){10-11}\cmidrule(lr){12-13}
        \multicolumn{1}{l|}{Defenses}
        & \multicolumn{1}{c}{MA} & \multicolumn{1}{c}{BA} 
        & \multicolumn{1}{c}{MA} & \multicolumn{1}{c}{BA} 
        & \multicolumn{1}{c}{MA} & \multicolumn{1}{c||}{BA} 
        & \multicolumn{1}{c}{MA} & \multicolumn{1}{c}{BA} 
        & \multicolumn{1}{c}{MA} & \multicolumn{1}{c}{BA} 
        & \multicolumn{1}{c}{MA} & \multicolumn{1}{c}{BA} \\
        \midrule
        FedAvg (*)  
        & 84.32 & 71.43 & 84.32 & 71.44 & 10.00 & 100.0 
        & 98.49 & 95.02 & 10.00 & 100.0 & 10.00 & 100.0 \\
        \cmidrule(lr){1-13}
        Krum
        & 81.73 &	\textbf{\textcolor{blue}{12.76}} &	79.54 &	18.78 &	81.58 &	\textbf{\textcolor{blue}{11.74}}
        & 95.51 &	98.88 &	95.52 &	89.96 &	95.83 &	96.74  \\
        Multi-krum 
        & \textbf{\textcolor{blue}{84.48}} &	67.35 &	\textbf{\textcolor{blue}{84.28}} &	67.38 &	10.00 &	100.0 
        & 98.38 &	94.87 &	98.36 &	96.65 &	10.00 &	100.0  \\
        FoolsGold
        & 84.27 &	68.88 &	84.08 &	66.33 &	\textbf{\textcolor{blue}{83.91}} &	71.93 
        & \textbf{\textcolor{red}{98.99}} &	75.13 &	\textbf{\textcolor{red}{98.99}} &	75.21 &	\textbf{\textcolor{red}{98.99}} &\textbf{\textcolor{blue}{81.13}} \\
        RFA
        & 82.98 &	88.27 &	82.89 &	88.23 &	72.01 &	95.41 
        & 95.33 &	97.91 &	95.29 &	98.01 &	97.17 &	96.88 \\
        RLR
        & 82.87&	15.31&	82.89&	\textbf{\textcolor{blue}{15.30}}&	83.32&	13.78 
        & 94.38 & \textbf{\textcolor{blue}{45.06}} &	94.33 &	\textbf{\textcolor{blue}{42.13}} &	96.33 &	85.81 \\
        FLAME
        & 83.60 &	71.43 &	83.60 &	72.42 &	82.95	& 76.53 
        & 98.24 &	96.97 &	98.38 &	95.98 &	98.38 &	96.26 \\
        \textbf{FedGrad}
        & \textbf{\textcolor{red}{84.62}} &	\textbf{\textcolor{red}{5.10}} &	\textbf{\textcolor{red}{85.10}} &	\textbf{\textcolor{red}{4.08}} &	\textbf{\textcolor{red}{85.37}} &	\textbf{\textcolor{red}{5.61}} 
        & \textbf{\textcolor{blue}{98.81}} &	\textbf{\textcolor{red}{7.98}} &	\textbf{\textcolor{blue}{98.83}} &	\textbf{\textcolor{red}{8.02}} &	\textbf{\textcolor{blue}{98.87}} &	\textbf{\textcolor{red}{7.96}} \\
        \bottomrule
        \end{tabular}
        }
        \captionof{table}{\footnotesize
        Backdoor Accuracy (BA) and Main Task Accuracy (MA) in percentages of FedGrad and baseline defense methods 
        within $1000$ rounds. 
        {The best and second best results are highlighted in the \textbf{\textcolor{red}{bold-red}} and \textcolor{blue}{\textbf{bold-blue}}. (*) \textit{no-defense} method.\label{table:accuracy}}}
   
\end{minipage}
\hspace{0.3cm}
\begin{minipage}{0.38\linewidth}
    	\centering
    	\small
    	\setlength\tabcolsep{4pt} 
    	\resizebox{\columnwidth}{!}{%
        \begin{tabular}{@{}l|l|ll|ll@{}}
        \toprule
        \multicolumn{1}{c|}{Trigger-based}
        & \multicolumn{1}{l|}{}
        & \multicolumn{2}{c}{FedAvg} 
        & \multicolumn{2}{c}{FedGrad}
        \\
        \cmidrule(lr){3-4}\cmidrule(lr){5-6}
        \multicolumn{1}{c|}{attack strategies}
        &\multicolumn{1}{c|}{Dataset}
        & \multicolumn{1}{c}{MA} & \multicolumn{1}{c}{BA} 
        & \multicolumn{1}{c}{MA} & \multicolumn{1}{c}{BA} \\
        \midrule
        \multirow{2}{*}{\begin{tabular}[]{@{}l@{}} Unconstrained \\attack\end{tabular}} 
        & MNIST & 99.07 & 99.78 & 98.94 & 0.14 \\
        & CIFAR-10 & 76.40 & 92.58 & 75.46 & 2.24 \\
        \midrule
        \multirow{2}{*}{\begin{tabular}[]{@{}l@{}} Constrain-and \\scale\end{tabular}} 
        & MNIST & 
99.07 &
99.60 &
98.97 &
0.12 \\
        & CIFAR-10 & 70.69 & 79.67 & 74.06 & 4.39 \\
        \midrule
        \multirow{2}{*}{\begin{tabular}[]{@{}l@{}}DBA\end{tabular}} 
        & MNIST & 99.07  & 100.0 & 98.90 & 0.10 
 \\
        & CIFAR-10 & 43.47 & 69.29 & 71.95 & 5.78 \\
        \bottomrule
        \end{tabular}
        }
        \captionof{table}{\footnotesize
        Backdoor Accuracy (BA) and Main Task Accuracy (MA) in percentages of FedGrad and  \textit{no-defense} method (FedAvg) against different trigger-based attack strategies within $400$ rounds.
        \label{table:dba_accuracy}}
   
\end{minipage}
\vspace{-0.2cm}
\end{figure*}
In this section, we evaluate the performance of FedGrad against various
backdoor attack scenario, including edge-case attacks~\cite{NEURIPS2020_b8ffa41d} and trigger-based backdoor attacks~\cite{pmlr-v108-bagdasaryan20a,DBA}.
We show that FedGrad is able to minimize the attack effect without degrading its learning performance compared with state-of-the-art FL methods including Krum/Multi-krum\cite{krum}, FoolsGold \cite{foolsgold}, RFA\cite{rfa}, RLR\cite{rlr}, and FLAME\cite{flame}. 
In the following, we first summarize our experimental settings and the evaluation metrics. We then report and compare the performance of FedGrad to the 
other defense methods. Moreover, we 
investigate the effectiveness of FedGrad's components and
the stability of FedGrad in different FL settings.

\subsection{Experimental Setup}

\textbf{Backdoor attack strategies.} We evaluate the performance of FedGrad under three edge-case based attack strategies proposed in \cite{NEURIPS2020_b8ffa41d}, which are black-box attack, projected gradient descent (PGD) attack, and PGD attack with model replacement (PGD + MR). For each strategy, we evaluate on two scenarios, e.g.,  LeNet~\cite{lenet} network on EMNIST~\cite{emnist} and VGG-9~\cite{vgg} on CIFAR-10~\cite{CIFAR-10} datasets. 
Additionally, we also evaluate FedGrad with a trigger-based backdoor attack, i.e., adding a small pattern to the inputs to backdoor the model, to prove the effectiveness of FedGrad against other backdoor attacks. We follow 
three strategies including (i) unconstrained attack \cite{Sun2019CanYR}, (ii) constrain-and-scale attack \cite{pmlr-v108-bagdasaryan20a} and distributed backdoor attack (DBA) \cite{DBA} under classification task on MNIST/CIFAR-10 with CNN/ResNet-18 \cite{resnet}.

\textbf{Adversary settings.} 
We establish a rigorous backdoor attack scenario by adjusting four parameters: the compromising ratio ($\epsilon$), the number of participants per communication round ($K$), the poisoned data rate ($PDR$) and non-IID degree ($\varphi$). In PGD attack, the projected frequency is set to 1 since there may be malicious clients in each training round. 
In trigger-based backdoor attacks, 
we use the same experimental setups in~\cite{DBA} with little modification as summarized in Table~\ref{tab:train_config_dba}
to align with the fixed-pool attack~\cite{NEURIPS2020_b8ffa41d}. Unless otherwise mentioned, our default settings are 25\% compromising ratio, non-IID level of $\varphi = 0.5$, \nguyen{poisoned data rate $PDR = 0.33/0.08/0.5$ and $K=10/20/30$ for CIFAR-10, MNIST, and EMNIST datasets, respectively. Other unmentioned parameters for reproducing are inherited from~\cite{NEURIPS2020_b8ffa41d,DBA}}.

\nguyen{\textbf{Hyperparameters.} 
\nguyen{We also follow the suggestion of the authors to set the robust threshold of RLR \cite{rlr} to $\left \lfloor K\cdot F+1 \right \rfloor$ 
and the estimated number of byzantine clients of Krum/Multi-Krum~\cite{krum} to $\left \lfloor \epsilon \cdot K \right \rfloor$.
For FedGrad, we set the dynamic filtering threshold of compromising score $\zeta=0.5$, threshold for trustworthy scores $\gamma=0.75$, instant trustworthy score for a malicious and benign client $\lambda_1 = 0.25$ and $\lambda_2 = 1.0$.} 

\textbf{Evaluation metrics.} To evaluate the effectiveness of the proposed approach, we leverage Main Accuracy (MA) and Backdoor Accuracy (BA), which are widely used in \cite{pmlr-v108-bagdasaryan20a,NEURIPS2020_b8ffa41d}. Moreover, we employ True Positive Rate (TPR) and False Positive Rate (FPR) to measure the efficiency of FedGrad \fix{in} detecting malicious clients.}
\begin{figure*}[!t]
\begin{minipage}{0.72\linewidth}
    \begin{minipage}{\linewidth}
        \centering
         \small          \includegraphics[width=1\textwidth]{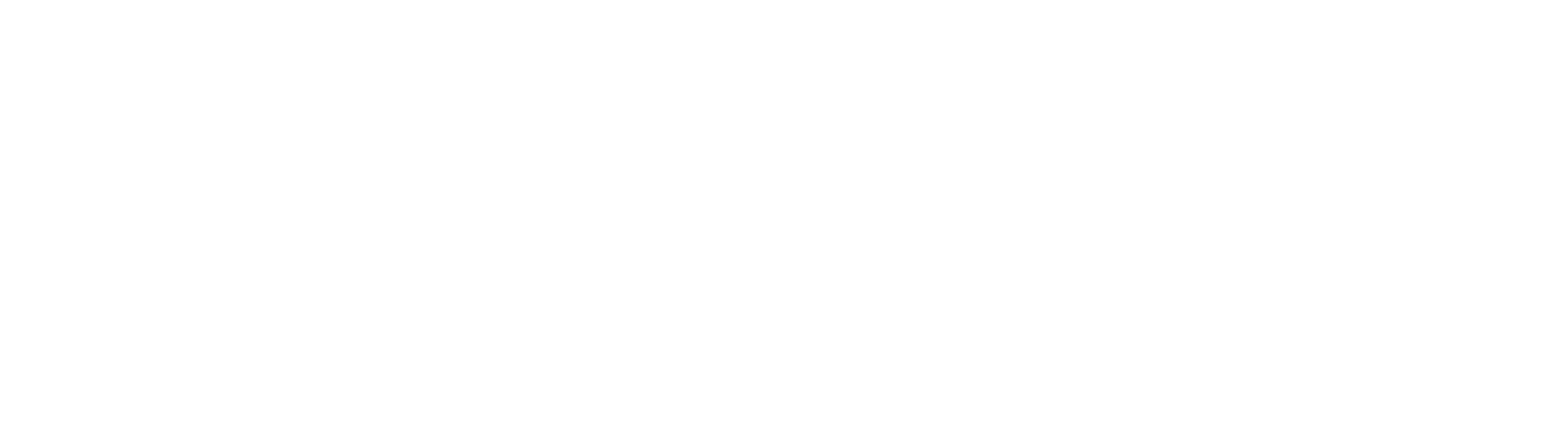}
          \\[-0.2cm]
         \caption{\footnotesize Backdoor accuracy vs. communication round under three attack strategies: Black-box, PGD, and PGD with model replacement attack (PGD + MR) on the CIFAR-10 dataset. {The results are plotted with the smoothed average of every $20$ round to have a better visualization.}
         \label{fig:backdoor_acc_result}}
    \end{minipage}
\end{minipage}
\hspace{0.3cm}
\begin{minipage}{0.26\linewidth}
    \begin{minipage}{\linewidth}
        \vspace{0.1cm}
    \centering
    \small
    \renewcommand{\arraystretch}{1.1}
    \centering
        \setlength\tabcolsep{3pt} 
        \resizebox{1\linewidth}{!}{
        \begin{tabular}{ l|cccc  }
         \hline 
         \textbf{Component} & \textbf{MA}    & \textbf{BA}    & \textbf{TPR}  & \textbf{FPR}  \\
         \hline
         \begin{tabular}[]{@{}l@{}} (1) Soft \\filtering\end{tabular} & 81.62 & 48.17  & 0.95 & 0.36 \\
         \hline
         \begin{tabular}[]{@{}l@{}} (2) Hard \\clustering\end{tabular} & 81.74  & 76.04 & 0.68 & 0.35 \\
         \hline
         (1) union (2) & 81.75 & 2.99  & 1.0  & 0.43 \\
         \hline
         \textbf{\textit{FedGrad}}
         & 82.13  & 2.54  & 1.0  & 0.3 \\
         \hline
        \end{tabular}
        }
        \captionof{table}{\label{tab:ablation_study} \footnotesize Effectiveness of components in FedGrad against the black-box edge-case backdoor attack on the CIFAR-10 dataset. TPR and FPR are average rate within $100$ communication rounds.}
\renewcommand{\arraystretch}{1}
    \end{minipage}
\end{minipage}
\vspace{-0.5cm}
\end{figure*}

\begin{figure*}[!t]
  \begin{minipage}{\linewidth}
        \centering
        \begin{minipage}{0.75\linewidth} 
             \centering
             \includegraphics[width=\columnwidth]{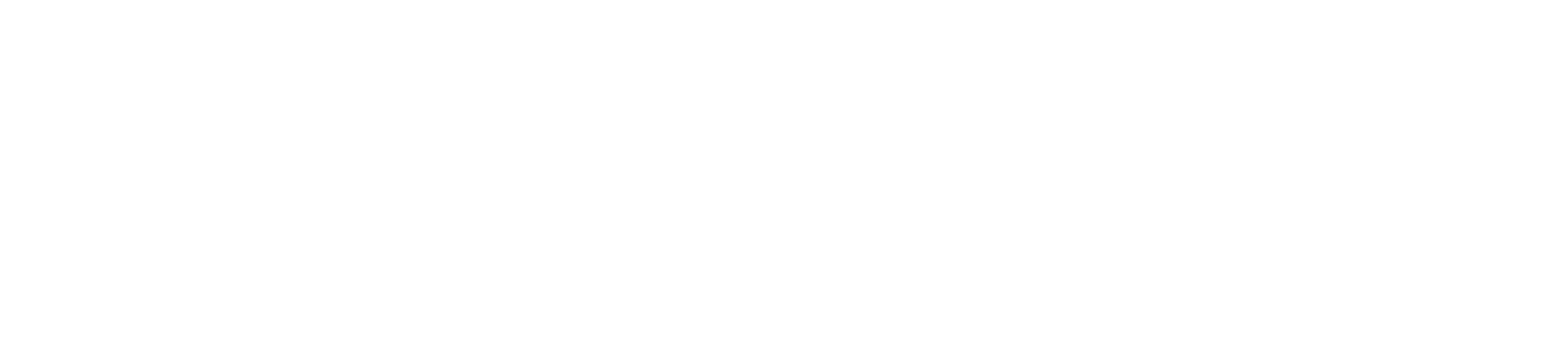}
              \caption{\footnotesize FedGrad against PGD 
              attack within 500 rounds in the EMNIST dataset when varying the compromising ratio (left), the poisoned data rate (middle) and the number of participants (right).}
              \label{fig:change_setting_acc}
        \end{minipage}
        \hfill
        \begin{minipage}{0.23\linewidth} 
            \centering
            \includegraphics[width=1\columnwidth]{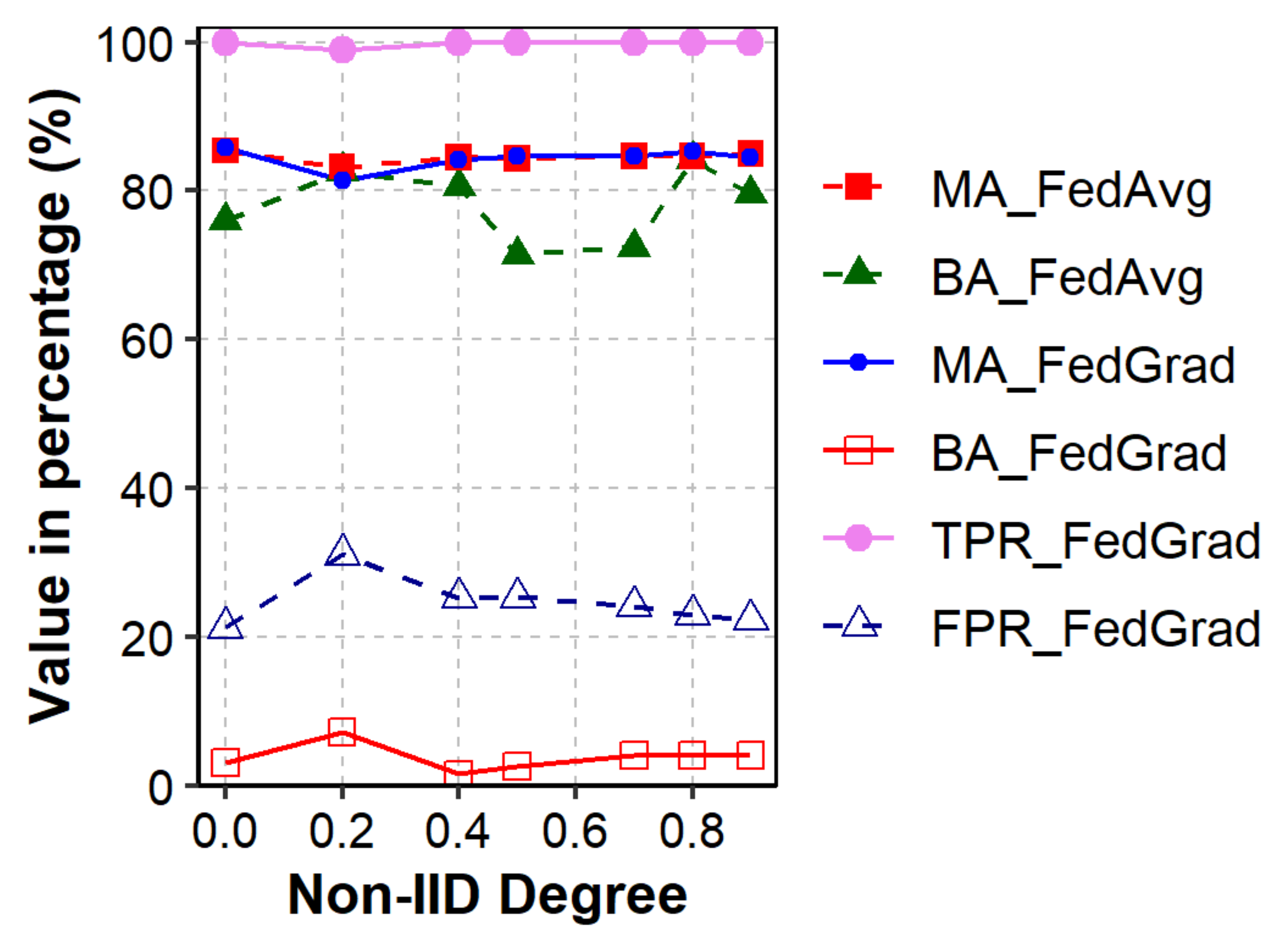} 
            \caption{\footnotesize Impact of non-IID level 
            (PGD attack, CIFAR-10).} 
            \label{fig:noniid-degree-result}
        \end{minipage}  
    \end{minipage}
    \vspace{-0.1cm}
\end{figure*}
\subsection{Defending Against Backdoor Attacks} In this section, we compare the MA and BA of FedGrad to the other defense methods to show the consistent effectiveness of FedGrad against different backdoor attack strategies.

\textbf{Edge-case backdoor attacks.} Table~\ref{table:accuracy} shows the best accuracy that each FL method reaches during training within $1000$ communication rounds.
The result shows that FedGrad outperforms other defense methods in all three attack strategies in terms of BA while maintaining a similar (or better) MA. 
On the CIFAR-10 dataset with a black-box attack, most defense methods fail to mitigate the effects of the backdoor on the aggregated model except RLR, Krum, and FedGrad. For example, other methods achieve high BAs ($65\%$), while those of RLR, Krum, and FedGrad are only $15.31$\%, $12.76$\%  and $5.10$\%, respectively.  
Compared with RLR and Krum in all investigated attack strategies, the proposed FedGrad achieves the lowest BA and outperforms RLR and Krum by around $1$\% and $4$\% on MA, respectively.
It is reported that DeepSight\cite{Rieger2022DeepSightMB}, the most relevant work to our approach, can also reduce the BA to $7.14\%$ with the MA of $80.54\%$ under the black-box attack (CIFAR-10). In this case, FedGrad outperforms DeepSight \footnote{We could not reproduce the result of DeepSight because there exists no public code at the time of writing this manuscript. 
} about $\approx 2\%$ in BA and $4\%$ in MA.
On the EMNIST dataset, all considered methods fail to defend against backdoor attacks except FedGrad. FedGrad is the only defense that can minimize the backdoor effect while not degrading the overall performance (BA $\approx 8$\% and MA $\approx 99$\%).

In more detail, Fig.~\ref{fig:backdoor_acc_result} presents the BA when the number of communication rounds changes on the CIFAR-10 dataset.  
RLR and RFA, the defense methods that do not focus on detecting and discarding adversarial clients, are not able to discard all the backdoor effects on the trained model. Defense methods utilizing the similarities between the whole dimension of local updates, i.e., FLAME, FoolsGold, and Krum, may fail to discriminate between benign and poisoning updates due to the convergence of local models compared with the global objectives. As a result, the performance of RLR and Krum gradually diminishes in later communication rounds.
\dung{As observed in Figure 6, the PGD attack degrades the performance of Krum at around the $640^{th}$ communication round, when the model starts converging. The PGD technique helps the poisoning models bypass the Krum method and cautiously be inserted into the global model when the model converges. This degradation phenomenon is also perceived in previous work~\cite{NEURIPS2020_b8ffa41d}.} 
In addition, the model replacement technique causes instability in RLR's performance. By contrast, FedGrad performs consistently in all examined attack strategies and gives the lowest BA during the training process. 
Specifically, FedGrad could detect the adversarial clients correctly, e.g., TPR $\approx 100\%$ on 
CIFAR-10 dataset (Table~\ref{tab:ablation_study}).

\textbf{Trigger-based backdoor attacks.}  
Table~\ref{table:dba_accuracy} proves the effectiveness of FedGrad against not just edge-case attacks, our main target in this work, but also trigger-based backdoor attacks. In all targeted 
strategies, FedGrad could minimize the effect of the backdoor attacks, e.g., BA $<$1\% in the case of MNIST, while not degrading the overall performance, i.e., less than 0.5\% of accuracy reduction when compared to those of FedAvg. 
Even under an intensive attack scenario, where FedAvg shows a significant \fix{decline} in MA, FedGrad still maintains the overall performance, e.g., 43.47\% and 71.95\% of MA for FedAvg and FedGrad
under the DBA attack.

\subsection{Effectiveness of FedGrad’s Components}
We investigate the contribution of each FedGrad component to the overall performance of FedGrad by independently performing defenses with each component used in FedGrad. Table~\ref{tab:ablation_study} reports the MA, BA, TPR, and FPR of each component when defending against backdoor tasks on the CIFAR-10 dataset. We observed that the second layer (\textit{By-class Ultimate Gradient-based Hard Filter}) works as a supplementary filter for the first layer (\textit{Ultimate Gradient-based Soft Filter}) to boost the TPR and decrease the BA significantly. Besides, the additional trustworthy filtering makes FedGrad provide a lower FPR. Hence, it can improve the MA by removing benign clients mislabeled as malicious while not degrading the effectiveness of detecting adversarial clients. 


\subsection{Stability Under Different Attack Settings}
We now demonstrate the stability of FedGrad by varying the three parameters, i.e., the number of malicious clients (compromising ratio), the number of participating clients, and the poisoned data rate. We compare FedGrad to other baseline defenses in terms of BA and MA (Fig.~\ref{fig:change_setting_acc}). 
In all settings, FedGrad outperforms all other defenses
and demonstrates its consistent performance.
Changing the attack setting causes a great impact on the performance of some defenses, i.e., RLR, FoolsGolds, and Krum. Meanwhile, the performance of RFA, MultiKrum, FLAME, and FedGrad is stable.
Especially,  FedGrad achieves the lowest BA in most cases (BA $\approx$ 8-10\%) while achieving a higher MA than those of FedAvg (no-defense method). It is worth noting that FedGrad, with its two-layer filtering mechanism, can maintain a high TPR in detecting malicious clients (as mentioned in Table 3). This observation also can be seen again in all the experiments shown in Fig.~\ref{fig:change_setting_acc}.
This result implies that our proposed mechanism for detecting and discarding malicious clients is stable and effectively mitigates the backdoor attack.
\subsection{Impact of Data Heterogeneity Degree}
Because FedGrad is primarily dependent on evaluating the differences between benign and malicious updates, the distribution of data among clients may impact our defense. We conduct experiments under the PGD attack \cite{NEURIPS2020_b8ffa41d} on the CIFAR-10 dataset with different degrees of non-IID distribution between clients' data.  We vary the degree of non-IID ($\varphi$) by adjusting the proportion of samples allocated to clients that belong to a specific class. In particular, we randomly divide the set of all clients into $10$ groups corresponding to $10$ classes in the CIFAR-10 dataset. A fraction of $\varphi$ samples of each class are assigned to group clients associated with this class, and the remaining samples are randomly distributed to others. As a result, a non-IID degree $\varphi$ is zero means the data is distributed completely IID (homogeneity), likewise, the data distribution is non-IID when $\varphi$ equals one.
As reported in Fig.~\ref{fig:noniid-degree-result}, FedGrad is able to detect all the adversarial models for even very non-IID scenarios, and it mitigates the effect of backdoors by maintaining a very low ($\approx 5\%$) in every case of non-IID level. Besides, FedGrad can erase the backdoor effect without a negative impact on overall performance. Specifically, the MA is maintained at almost the same level as without defense. Even in the case of a low non-IID level of homogeneous data distribution (IID), FedGrad performs well with a maximum TPR and an acceptable FPR. 


\section{Discussion}
\label{sec:discussion}
\subsection{\nguyen{Computation Overhead of FedGrad}} 
Our goal in this study is to design an efficient approach for a synchronous FL system in which the server waits for all clients to complete their local training before clustering the participating clients and computing the new global model at each communication round.
A significant \fix{amount of} computation time on the server would prolong the synchronization latency. 
We thus estimate the computation time at the server (averaged within 150 communications rounds) under a black-box attack on the CIFAR-10 dataset. 
Notice that we simulate FL on a computer with an Intel Xeon Gold CPU and an NVIDIA V100 GPU.

Fig.~\ref{fig:compute_overhead} shows that the computation time of FedGrad is lower than those of other defense methods except for the RLR method. For example, the computation time at the server of FedGrad is about $8.2\times$ and $1.2\times$ lower than those of RFA and Krum-based methods.
Interestingly, the time of the trustworthy filtering component is trivial, while hard filtering and soft filtering contribute most of the computation time of FedGrad, i.e., $54$\% and $45$\%, respectively. It is a reasonable result because the computation complexity of the similarity computation in each communication round of the hard clustering component is $O(K^2)$ where $K$ is the number of clients participating in each round. To speed up such an operation in the server, computing the similarity by batch instead of by pair of clients could be helpful. That is, we compute the similarity of a given client $i$ with $b$ other clients ($b \le K$) at the same time and utilize the high parallelism of multi-threading on GPU. In short, our FedGrad requires an acceptable computation time at the server, which is still practiced in real-world applications.
\subsection{\nguyen{Convergence Time of Client Similarity Determination}}
One of the core findings and technologies in this work is to use the by-class ultimate gradient to filter and detect malicious clients (see Section \ref{subsec:hardfilter}).
For each training round, we need to calculate the similarity between $K$ clients, resulting in a computational complexity of $K^2$. Therefore, if the FL training process takes $m$ rounds to converge, the overall computational complexity will be $m \times K^2$.
To this end, we may reduce this cost by determining the similarity transitively. 
Specifically, the similarity of two arbitrary clients, $C_i$ and $C_j$ can be estimated via their similarities with other clients. This way, we can shorten the number of training rounds needed for the similarity to converge.
We notice that cosine similarity possesses a transitive characteristic, which is reflected by the theorem in~\cite{schubert2021triangle}. Based on this, we have the following theorem.

\begin{theorem}
\label{pro:similarity}
Let $n$ be the total number of clients and $K$ be the number of clients participating in a communication round. Let $\delta$ be the expected number of communication rounds needed to estimate the similarity of all client pairs.
Then, $\delta \leq 1 + \sum_{i=K}^{n-1}\frac{\binom{n}{k}}{\binom{n}{K}-\binom{i}{k}}$.
\end{theorem}

\begin{figure}[tb]
    \centering
    \begin{minipage}{0.42\linewidth}
        \includegraphics[width=\linewidth, trim=0 0.5cm 0 0cm]{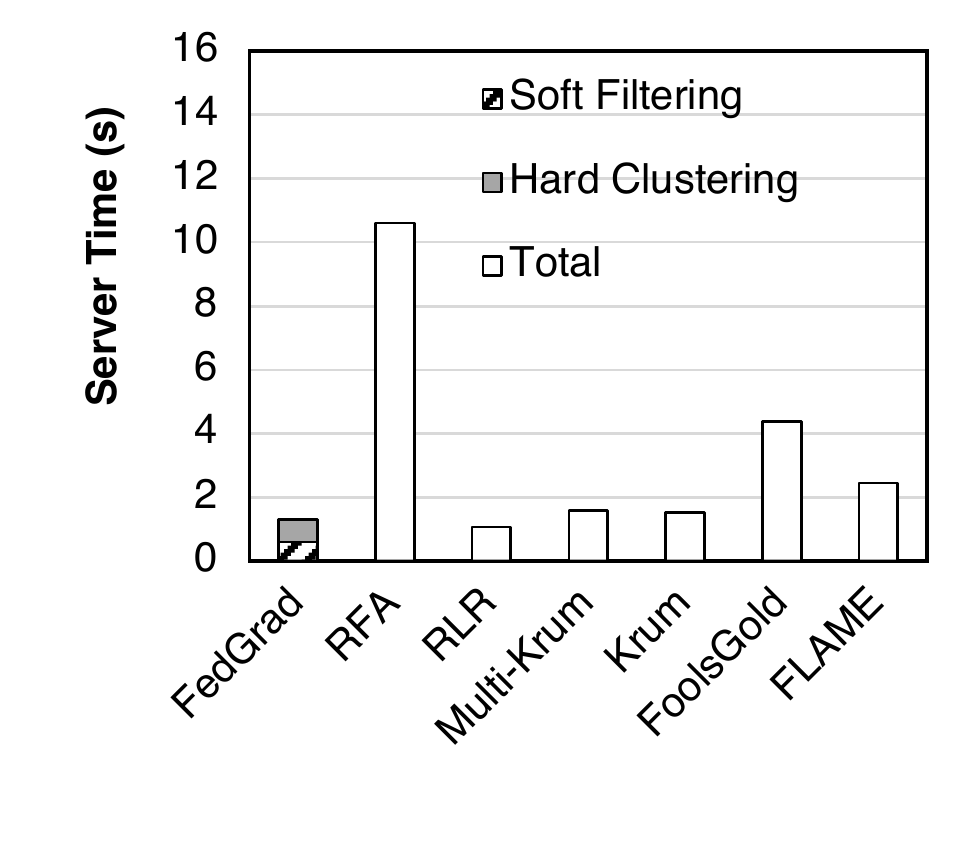}
         \caption{\small{Average server computation time.
         \label{fig:compute_overhead}
         }}
    \end{minipage}
    \hfill
    \begin{minipage}{0.54\linewidth}
        \includegraphics[width=\linewidth, trim=0 3cm 0 3cm, clip]{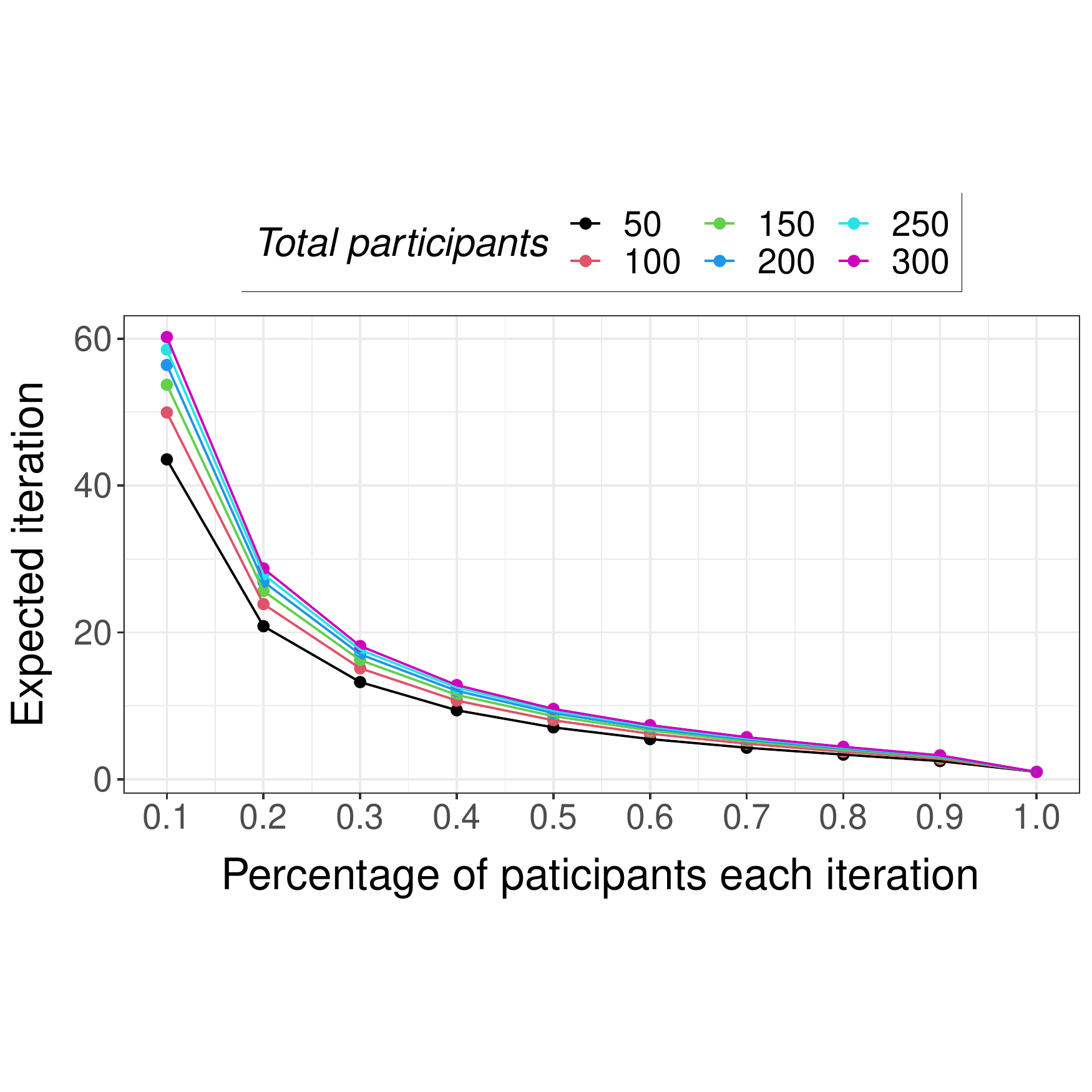}
        \caption{\small{Estimations of required iterations for convergence of similarity. \label{fig:convergence}}}
    \end{minipage}
\end{figure}

\nguyen{
We conducted extensive experiments to estimate the convergence of client similarity of FedGrad with different settings, i.e., the total number of clients $n$ ranges from $50$ to $350$ while the participation rates $\frac{K}{n}$ are fixed at $0.2, 0.4, 0.6, 0.8$ and $1.0$. It is clearly shown in Figure~\ref{fig:convergence} that the expected number of training rounds required for convergence of similarity is trivial concerning the number of clients. For instance, when there are $n = 100$ clients in total, the similarity is expected to converge after roughly $25$ iterations (or communication rounds) at a participation rate of $0.2$ and after about $50$ communication rounds at a participation rate of $0.1$.
}
\subsection{\nguyen{FedGrad under different attack scenarios and assumptions}}
\nguyen{All the experiments in section~\ref{sec:evaluation} are based on the common attack scenarios~\cite{NEURIPS2020_b8ffa41d,Sun2019CanYR} in which: (1) the rate of malicious clients cannot exceed $50$\% in real-world settings; and (2) the data distributions and the amount of perturbation caused by malicious clients are the same. The following sections analyze FedGrad's behavior under more severe attack scenarios.}

\nguyen{Firstly, we show that FedGrad can achieve a good performance under \textbf{a high ratio of malicious clients, e.g., 70\%.} by adjusting the dynamic threshold of the soft filter. Specifically, in previous experiments, When the rate of malicious clients smaller than $50$\%, we set the dynamic threshold of the soft filter to $\epsilon_1 = min (\zeta, \text{median } cs_i)$. 
For stricter filter, i.e., the malicious ratio reaches 70\%, the threshold could be adaptively changed, e.g., set to $0.35\times \epsilon_1$.
Additional experiments with 70\% attacker show that such little modification can help FedGrad effectively mitigate the backdoor attack (BA$\sim$10\%). By contrast, Krum let malicious clients bypass, and RLR significantly degrades the MA compared to FedGrad (i.e., FedGrad's MA $\sim$ 98.5\% and RLR's MA $\sim$ 91.5\%). It's worth noting that even under the benign setting (no malicious clients) the BA is around 10\%.
}

\nguyen{We also run an additional experiment where \textbf{malicious clients have varying poisoned data rate (PDR).} Specifically, under the presence of $25$\% of compromised clients, we set up an attack with $80$\% of malicious clients using PDR of $15$\% and the remaining $20$\% using the PDR of $50$\% with the EMNIST dataset. FedGrad can still maintain a very low BA ($\sim 5\%$) without compromising overall performance (MA $\sim 99.4\%$).}

\nguyen{Finally, the most favorable attack scenario is that the attacker adds a set $\mathcal{D}_{p}$ that describes the backdoor task (i.e., southwest airplane images) into the benign training samples generated from the true distribution, which is aligned with the previous works~\cite{pmlr-v108-bagdasaryan20a,NEURIPS2020_b8ffa41d}. To make the attack stealthier, each compromised client itself should achieve sufficient accuracy on the main task by maintaining a true distribution of benign data together with $\mathcal{D}_{p}$. Therefore, the most practical and non-effortless strategy for the adversary is to sample data from this true distribution to replicate each malicious client.We now consider the attack scenario where \textbf{malicious clients have different data distribution}. We conducted extra experiments with the edge-case backdoor attack with the EMNIST dataset, where the data distribution of malicious clients is also non-IID. In this scenario, FedGrad successfully mitigates the backdoor and maintains BA at $\sim 5\%$ while maintaining MA $\sim 98.8\%$.
}

\section{Conclusion}
We proposed in this paper a novel backdoor-resistant defense named FedGrad by investigating ultimate gradients of clients' local updates. FedGrad leverages a two-layer filtering mechanism that can accurately identify malicious clients and remove them from the aggregation process.  
We conducted a comprehensive experiment set to evaluate the effectiveness of FedGrad on various datasets and attack scenarios. The results indicated that FedGrad outperforms state-of-the-art FL approaches in defending against edge-case backdoor attacks and mitigating perfectly other common backdoor attack types, especially when the number of compromised clients is significant and the data distribution across clients is highly non-IID. Our future work will be devoted to handling backdoor attacks in more realistic scenarios, such as large-scale networks and hierarchy networks.

\section{Acknowledgments}
This work was funded by Vingroup Joint Stock Company (Vingroup JSC), Vingroup, and supported by Vingroup Innovation Foundation (VINIF) under project code VINIF.2021.DA00128. This paper is based on results obtained from a project, JPNP20006, commissioned by the New Energy and Industrial Technology Development Organization (NEDO). 	\\

\bibliographystyle{unsrtnat}
\bibliography{references}  






\end{document}